\documentclass{article}

\usepackage{resources/PRIMEarxiv}

\usepackage[utf8]{inputenc} 
\usepackage[T1]{fontenc}    
\usepackage{hyperref}       
\usepackage{url}            
\usepackage{booktabs}       
\usepackage{amsfonts}       
\usepackage{nicefrac}       
\usepackage{microtype}      
\usepackage{lipsum}
\usepackage{fancyhdr}       
\usepackage{graphicx}       
\graphicspath{{media/}}     

\usepackage{framed,multirow}
\usepackage{amssymb,amsopn}
\usepackage{amsmath}
\usepackage{import}
\usepackage{graphicx}
\usepackage{latexsym}
\usepackage{caption}
\usepackage{subcaption}
\usepackage{color,soul}
\usepackage{cleveref}

\usepackage{todonotes}

\usepackage{url}
\usepackage{xcolor}
\definecolor{newcolor}{rgb}{.8,.349,.1}
\usepackage{array}

\usepackage{lipsum}
\usepackage{amsfonts}
\usepackage{graphicx}
\usepackage{epstopdf}
\usepackage{algorithmic}
\ifpdf
  \DeclareGraphicsExtensions{.eps,.pdf,.png,.jpg}
\else
  \DeclareGraphicsExtensions{.eps}
\fi

\DeclareMathAlphabet{\bi}{OML}{cmm}{b}{it}

\newcommand{\rme}{\mathrm{e}}
\newcommand{\rmi}{\mathrm{i}}
\newcommand{\rmd}{\mathrm{d}}

\newcommand{\IKM}{\mbox{${\mathcal I}^{\scriptscriptstyle{\rm KM}}$}}

\newcommand{\xb}{\mbox{$\vec{\bi{x}}$}}
\newcommand{\yb}{\mbox{$\vec{\bi{y}}$}}

\newcommand{\ybs}{\mbox{$\vec{\bi{y}}^{\,\rm s}$}}

\newcommand{\w}{\mbox{$\omega$}}
\newcommand{\W}{\mbox{$\Omega$}}

\newcommand{\ds}{\displaystyle}

\title{A physically-informed Deep-Learning approach for locating sources in a waveguide\thanks{Corresponding author: AK (adarkahana@gmail.com). AK and SP contributed equally to this work. The research of DB and SP is supported by Israel Science Foundation grant 1793/20 and by Lower Saxony-Israel collaboration grant from the Volkswagen Foundation.}
}

\author{
  Adar Kahana, Symeon Papadimitropoulos, Eli Turkel, Dmitry Batenkov \\
  Department of Applied Mathematics \\
  Tel Aviv University \\
  Tel Aviv 69978, Israel \\
}

\begin{document}
\maketitle

\begin{abstract}
Inverse source problems are central to many applications in acoustics, geophysics, non-destructive testing, and more. Traditional imaging methods suffer from the resolution limit, preventing distinction of sources separated by less than the emitted wavelength. In this work we propose a method based on physically-informed neural-networks for solving the source refocusing problem, constructing a novel loss term which promotes super-resolving capabilities of the network and is based on the physics of wave propagation. We demonstrate the approach in the setup of imaging an a-priori unknown number of point sources in a two-dimensional rectangular waveguide from measurements of wavefield recordings along a vertical cross-section. The results show the ability of the method to approximate the locations of sources with high accuracy, even when placed close to each other.
\end{abstract}


\section{Introduction}


A large class of inverse problems in imaging aims at recovering locations of sources of waves from sensor measurements of the wavefield radiated by these sources. Many applications for locating sources exist in 
the literature, in various fields such as acoustics, geophysics,  non-destructive evaluation and more 
\cite{albocher, barbone,InvPro,InvPro3,InvPro4,InvPro5, allen,baer,kup,sleeman}.

These ``inverse source'' problems are usually ill-posed. The incomplete data provided only by a few sensors makes the solution very sensitive to that data. In addition, traditional imaging methods such as Kirchhoff migration suffer from the so-called resolution limit,  when close-by sources cannot be distinguished from each other in the image due to the nonzero width of the Green's function. Various \emph{super-resolution} techniques can in principle overcome these limitations, however at the expense of extreme sensitivity to noise in the data and highly nontrivial mathematical theory, which is currently applicable only in a limited number of cases (see \Cref{sec:Formulation}).

%
%

The use of machine-learning (ML) for solving inverse problems is spreading fast within the 
scientific community \cite{WaveGuide,pinns,zhu,arridge_solving_2019}. According to this paradigm, many PDE-based inverse problems of the form $A(x)\approx y$,  including the one in this paper, 
can be formulated as data-driven problems, i.e. searching for a general neural network model ${\cal NN}_{\theta}$  whose weights $\theta$ are learned by fitting a vector of model parameters $\{x_i\}$ to the corresponding measurements $y_i\approx A(x_i)$, requiring that ${\cal NN}_{\theta}(y_i) \approx x_i$ for all $\{(x_i,y_i)\}$ in a large training set. The 
advances in computation capabilities and ease of use of the ML libraries have opened doors for many 
researchers to investigate ML based inverse solutions corresponding to various forward models $A$. In particular, ML-based inversion techniques can be used with highly nonlinear $A$, are relatively robust to perturbations, and can provide real-time inference (in contrast with traditional  optimization-based methods where the solution to a single problem instance \mbox{$A(x^*) \approx y^*$} may require substantial computational resources).

ML-based methods also have their drawbacks. In a classical ML approach to inverse problems, the structure of the mapping ${\cal NN}_{\theta}$ is very general and independent of the governing PDE. Therefore, ML methods may severely underperform when tested on a slightly different task
than the one they were trained for. For example, using a model that was trained to find point sources will struggle when injecting data 
produced by using an extended source.

Deep-learning (DL) refers to a subset of ML methods, where the function ${\cal NN}_{\theta}$ is composed of a very large number of layers (which makes the network ``deep''), where each layer connects to the next by a pointwise, usually nonlinear, transformation. While the study of DL is an active area of research, it is frequently difficult to predict whether a DL approach would be superior to other ML methods to solve a particular inverse problem. DL architectures have been shown to successfully approximate nonlinear high-dimensional mappings, thereby becoming the ``go-to'' method for data-driven inverse problems.

An emerging class of approaches is the \emph{Physics-Informed Machine Learning}, which explicitly utilize some knowledge of the physics of the problem in order to design more robust and efficient solution methods, for both the forward and the inverse problems. Usually, the physical knowledge is introduced by specifically constructed additional loss terms, subsequently leveraging the powerful automatic differentiation \cite{autodiff} capabilities of modern ML tools \cite{karniadakis_physics-informed_2021}.

\subsection{Contributions}

We propose a method based on deep neural networks for solving the source refocusing problem. The contributions are twofold: first, we use tools from computer vision to construct a base DL architecture which can predict the locations of the sources in a pixel-wise fashion. Second, we construct a novel physically-informed (PI) loss term which promotes super-resolving capabilities of the network and is based on the physics of wave propagation. We demonstrate the approach in the setup of imaging an a-priori unknown number of point sources in a two-dimensional rectangular waveguide from measurements of wavefield recordings along a vertical cross-section. The network is trained on different configurations of sources, by solving the Helmholtz equation corresponding to these configurations with appropriate boundary conditions.

The results show the ability of the method to approximate the locations of sources with high accuracy, even when placed close to each other, thereby overcoming the classical resolution limit. Furthermore, the addition of the PI loss term dramatically improves the accuracy of the predictions and also helps the training process to converge more rapidly.

\section{Formulation of the problem} \label{sec:Formulation}

\begin{figure}[ht]
\begin{center}
\includegraphics[width=0.7\textwidth]{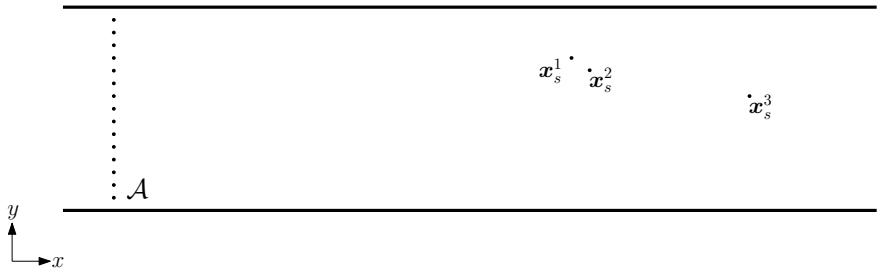}
\caption{A two-dimensional infinite waveguide, containing a vertical array $\mathcal{A}$ and three point sources $x_s^1, x_s^2$ and $x_s^3$.}
\label{fig:wg_setup}
\end{center}
\end{figure}

We consider the problem of locating multiple sources in an infinite waveguide with horizontal 
boundaries, see \Cref{fig:wg_setup}. We consider a Cartesian coordinate system $(x,y)$, 
with  $x$ denoting the main direction of propagation and $y$ the cross-range direction. 
We consider a homogeneous waveguide $\W$ with a constant wave speed $c_0$, and homogeneous Dirichlet boundary conditions on both the top and bottom boundaries. Wave propagation inside the waveguide is governed by the Helmholtz equation
\begin{equation} \label{eq:HE}
-\Delta p (\w,\xb) - k^2 p(\w,\xb) = g(\w,\xb), \xb \in \W,
\end{equation}
where $\w = 2 \pi f$ is the angular frequency and $k = \w / c_0$ is the wavenumber.

Let $G(\xb,\xb_s;\omega)$ denote the Green's function  for the Helmholtz operator and the associated boundary conditions, due to a point source located at $\xb_s = (x_s,y_s) \in \Omega$. For a single frequency $\omega$, $G(\xb,\xb_s;\omega)$ is the solution of 
\begin{equation}   \label{eq:Green1}
   -\Delta G(\xb,\xb_s;\omega) - k^2 G(\xb,\xb_s;\omega) = \delta(\xb-\xb_s).
\end{equation} 

Finally, let $(\mu_n,Y_n)$ be the eigenvalues and corresponding orthonormal eigenfunctions of the following vertical eigenvalue  problem:
\begin{equation} \label{eq:VE}
Y''(y) + \mu Y(y) = 0, \qquad Y(0) = Y(D) = 0.
\end{equation}
Henceforth we assume that there exists an index $M$ such that the constant wavenumber $k$  satisfies:
\begin{equation*}
\mu_M < k^2 < \mu_{M+1}.
\end{equation*}
Thus, $M$ is the number of \textit{ propagating modes} in $\Omega$.
We also denote the horizontal wavenumbers in $\Omega$ by
\begin{equation}    \label{eq:betas}
  \beta_n = \left\{\begin{array}{ll}
                  \sqrt{k^2 - \mu_n},      & 1 < n < M,\\
                  \rmi \sqrt{\mu_n - k^2}, & n> M+1 .\end{array}\right.
\end{equation}
In the case of the homogeneous infinite waveguide, the Green's function may be written analytically as

\begin{equation} \label{eq:greensf}
\ds G(\xb,\xb_s;\w) = \sqrt{\frac{2}{D}} \sum_{n=1}^{\infty} \frac{1}{\beta_n} \rme^{\rmi \beta_n |x-x_s|} \sin (\sqrt{\mu_n} y) \sin (\sqrt{\mu_n} y_s),
\end{equation}
where $\mu_n = n^2 \pi ^2 /D^2$.

\subsection{Array imaging setup}
 We assume that inside the waveguide, there exist $N_s$ point sources, each located at $\xb_i = 
 (x_i,y_i),~i=1,2, \ldots,N_s$, as well as a vertical array $\mathcal{A}$, consisting of $N_r$ 
 receivers,  that spans the entire vertical cross-section of the waveguide, as shown in 
 \Cref{fig:wg_setup}. The  acoustic pressure field generated by the sources is recorded on the array and 
 stored in the  array response matrix $\Pi$. In this case the array response matrix $\Pi$ at 
 frequency $\w$ reduces to a  $N_r \times 1$ vector, whose $r-$th component contains the Green's function due to each source $\xb_i $, evaluated at receiver $\xb_r$, i.e.,
 
 \begin{equation}\label{eq:response-matrix}
 \Pi(\xb_r;\w) = \sum_{i=1}^{N_s} G(\xb_r,\xb_i;\w).
 \end{equation}

\subsection{Super-resolution}
 
\begin{figure}[ht]
\begin{center}
\includegraphics[width=0.6\textwidth,clip]{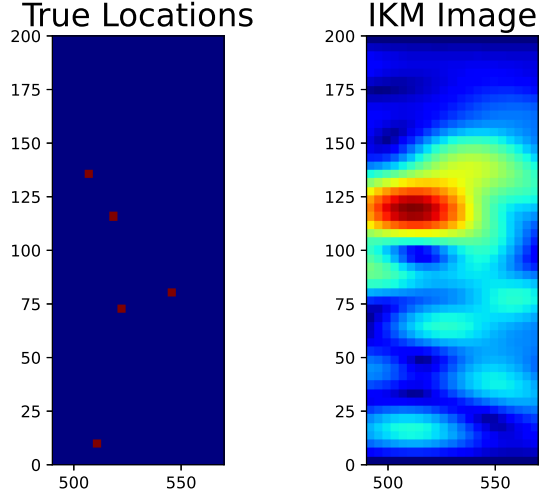}
\caption{Left: True locations (left) and $\IKM$ image (right) of five point sources located at $\xb_1 = (507,135)$~m, $\xb_2 = 
(519,116)$~m, $\xb_3 = (523,73)$~m, $\xb_4 =(546,80)$~m and $\xb_5 = (511, 10)$~m, for $f_c = 32.0625$~Hz, and $B = 0.4 f_c$ .}
\label{fig:IKM_example}
\end{center}
\end{figure}

Traditional imaging methods are based on \emph{imaging functionals} that back-propagate the response matrix to a search domain $\mathcal{S}$. These imaging functionals are designed such that when computed and graphically displayed, they form an image that  exhibits peaks at the location of the sources. One of the most well known such functional, is the Kirchhoff Migration (KM) imaging functional \cite{B_12}, given by 
 
 \begin{equation} \label{eq:IKM}
 \IKM(\ybs) = \sum_{\w} \sum_{n=1}^{N_r} \overline{\Pi(\xb_r;w)}G(\xb_r,\ybs;\w),
 \end{equation}
where $\ybs$ is a point in the search domain $\mathcal{S}$. An example of a K-M image can be seen in 
\Cref{fig:IKM_example}, where there are five point sources, located at $\xb_1 = (507,135)$~m, $\xb_2 = 
(519,116)$~m, $\xb_3 = (523,73)$~m, $\xb_4 =(546,80)$~m and $\xb_5 = (511, 10)$~m. The search domain is $\mathcal{S} = [490, 570] \times [0 200]$~m, and we use frequencies around a central frequency $f_c = 32.0625$~Hz, with a bandwidth $B = 0.4 f_c$.

From the generated image, we observe that the imaging functional exhibits local maxima at the locations of the sources, but it struggles in discerning between the ones that are placed close to one another. This is a well-known 
limitation of  such imaging methods, and is quantified by what is called the `resolution' of the imaging 
functional.  Performing a resolution analysis for any imaging functional, where we image a single point 
target, gives us an estimate of how close two targets can be in order for them to be well separated in 
the generated image. When using traditional imaging methods, the best resolution one can achieve is the \textit{Rayleigh 
limit} of $\lambda/2$, where $\lambda$ is the wavelength of the frequency used to create the image. In 
the case of multiple frequency imaging, as the one shown in \Cref{fig:IKM_example}, then resolution is tied to the wavelength of the central frequency $f_c$\cite{B_12}.

Linear imaging functionals such as \eqref{eq:IKM} effectively regularize the ill-posed inverse source problem by implicitly assuming that the imaged object is a smooth low-pass function. \emph{Super-resolution} (SR) is a body of mathematical and practical techniques which seek to overcome this limitation, by imposing different kinds of priors on the unknown object (in this context, the source distribution $g(\omega,\vec{x})$ in \eqref{eq:HE}). Classical SR methods assume small space-time  object extent, in which case analytic continuation or singular function expansions can be used to extract information from the evanescent modes (recall \eqref{eq:greensf}). However in such cases the number of degrees of freedom recovered beyond the Rayleigh limit scales only logarithmically with the signal-to-noise ratio \cite{batenkov_stable_2019, bertero_iii_1996, de_villiers_limits_2016, lindberg_mathematical_2012}. In recent years the \emph{sparse modeling} methods have  gained much popularity, both in general inverse problems \cite{daubechies_sparsity-enforcing_2016} and more specifically in SR \cite{candes_towards_2014, batenkov_super-resolution_2021}. These methods offer potentially much better accuracy and stability, at the expense of separation conditions, nonlinear reconstruction algorithms and complicated theoretical guarantees. Unfortunately, rigorous sparse SR techniques are nontrivial to generalize to wave-based imaging, and we are not aware of a theoretically solid and robust method for SR imaging even in this simple setup. 


\section{Deep Learning approach}
In this section we describe the deep learning based method for the solution of the imaging problem elaborated in the previous section. In Section \ref{sub:data-driven-form}, we formulate the mathematical problem as a data-driven problem. In Section \ref{Sec:DLF} we describe the DL architecture in detail, adding the physically-informed loss term in Section \ref{sub:pi-loss}. 

\subsection{Data-driven formulation}\label{sub:data-driven-form}

In this section, we formulate the source refocusing problem as a data-driven problem. We create different 
scenarios that vary in the number of sources placed in the domain, as well as their locations. 
Each scenario is called a sample, and the number of sources and their locations in each sample are 
arbitrary.  For each sample, a response matrix is created, as described in \Cref{sec:Formulation}, which 
will be the input of the network. In more detail, the input for the network is the vector of measurements recorded in the sensors, per sample. Therefore, each of the $N_{samples}$ 
samples has the recorded data in each of the $N_r$ receivers $\{\xb_r\}_{r=1}^{N_r}$ for each of the $N_f$ frequencies $\{\w_j\}_{j=1}^{N_f}$:
\begin{equation}\label{eq:exact-data}
d_q = \{\Pi(\xb_r;\w_j;q)\}_{r=1,\dots,N_r}^{j=1,\dots,N_f},\quad  \Pi(\xb_r;\w;q) = \sum_{i=1}^{N_s(q)} G(\xb_r,\xb_{i,q};\w),
\end{equation}
where each source configuration $q=1,\dots,N_{samples}$ consists of $N_s(q)$ sources at locations $\xb_{i,q}$.
The size  of the input is therefore $ N_r\times N_f$ for each sample.

\newcommand{\PL}{\ensuremath{\mathfrak{P}}}
\newcommand{\IM}{\ensuremath{\mathcal{I}}}

Given an input $d\in\mathbb{C}^{N_r\times N_f}$, we would like the network to predict the source locations $\{\xb_i\}$. One possibility is to define the network output to contain the exact coordinates of these sources. However, in this case we have to fix the number of sources $N_s(q)=\textrm{const}$, since the network must have a constant sized output. We propose an alternative method, where the network should predict \emph{an image} $\IM=\IM_q$ of dimension $N_x\times N_y$ corresponding to the source distribution $g$ as in \mbox{\eqref{eq:HE}}, such that
\begin{equation}\label{eq:plateau-image-def}
\IM_q = \bigvee_{i=1}^{N_s(q)} \PL(\xb_{i,q})
\end{equation}
where $\PL(\xb)$ stands for an $N_p \times N_p$ \emph{binary plateau} centered at the location of the 
source a $\xb$, with the value $1$ inside the plateau region and $0$ everywhere else, and $\vee$ is the pixel-wise binary OR operation. These binary images are precisely the associated \emph{labels}, being the desired outputs of the model that we use in the training phase (when training, the model fits input data to output labels).

The size of the output is therefore $N_x\times N_y$ for each sample. This method is popular in image segmentation problems and as we show later, turned out to be robust and effective in this context as well. On the flip side, we need to apply a post-processing method to infer the source locations from the predicted plateau images, as described in \mbox{\Cref{eval_method}}.

Choosing the plateau size $N_p$ turns out to be nontrivial as it incurs a tradeoff between stability and resolution. On the one hand, using a small number such as $N_p = 1$ or $N_p = 2$ creates sparse output images. From the experiments we observe that in those cases the network converges to the $0$ solution, in the sense that all the outputs of the network were $0$ images. This is due to the loss values being  very low in that case (the $0$ solution is very close to the sparse output in terms of the loss function). Thus, the network returns the $0$ solution as a local minimum. On the other hand, when using large plateaus we obtain lower resolution in the sources refocusing, since it is harder to distinctly identify the sources due to a large overlap. In practice we choose $N_p$ by starting with $N_p=1$ and gradually increasing $N_p$ until the network no longer converges to the zero solution.

\subsection{Deep-learning network architecture} \label{Sec:DLF}

\begin{figure}[ht]
\begin{center}
\includegraphics[width=0.85\textwidth]{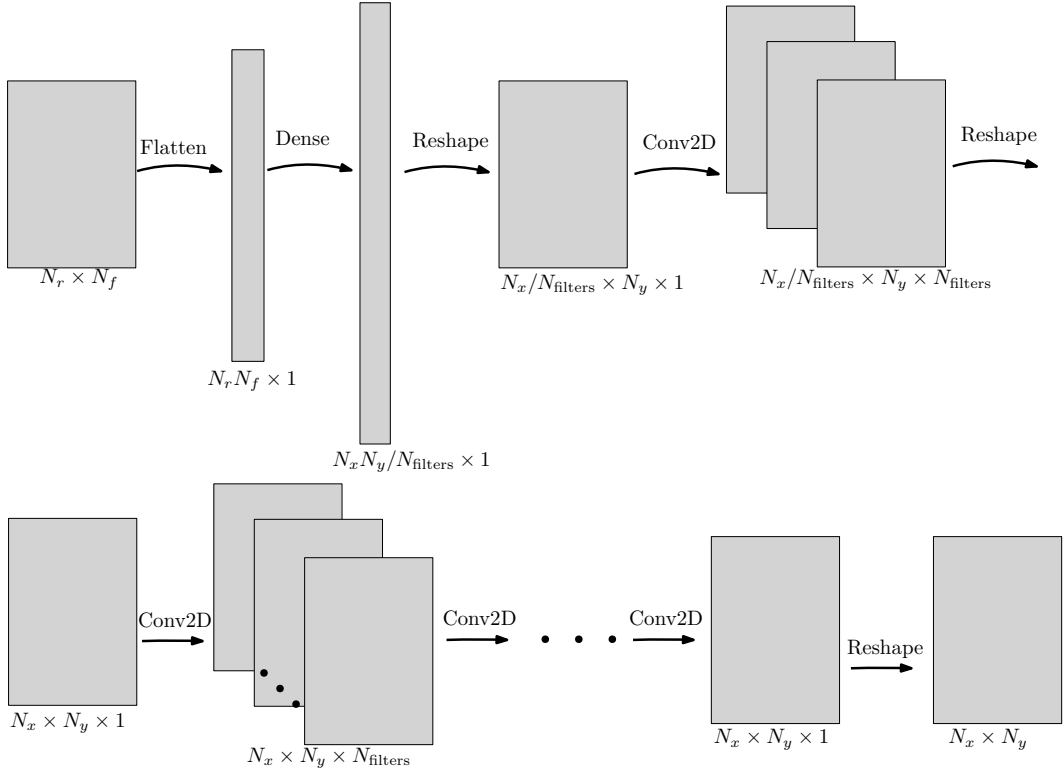}
\caption{The architecture of the proposed Deep Learning network.}
\label{fig:network_architecture}
\end{center}
\end{figure}

As described in the previous section, the predictor network ${\cal NN}$ is a mapping
\begin{align*}
{\cal NN}_{\theta}: \mathbb{C}^{N_r\times N_f} &\to \left[0,1\right]^{N_x\times N_y},\\
d &\mapsto \tilde{\IM}(d;\theta),
\end{align*}
where $d$ is the data image \eqref{eq:exact-data}, $\tilde{\IM}$ is the approximate binary plateau image (in fact, an image of probabilities, see below), and $\theta$ is a vector of \emph{weights} whose values are found by the learning (training) process to be described below. The architecture of the network is schematically shown in \Cref{fig:network_architecture}.

The network consists of a fully-connected (dense) layer which captures the global connections in 
the data, followed by a series of convolutional layers whose purpose is to capture local 
patterns, connected by the appropriate stacking and reshaping operations. We expect these local patterns to exist since two sources that are very close to each other will produce fields that are relatively similar, while the fields produced by sources that are far apart from each other will differ significantly. Similarly, the fields recorded by two close-by receivers will share a lot of characteristics, while if we, for example, compare recordings from two receivers on opposite ends of the array, we will find few to no similarities between them.

 Effectively, the 
original 2D spatio-frequency image is converted to a 2D spatial output, and the convolutional 
filters facilitate the additional spatial dimension. In \Cref{fig:network_architecture}, 
between the first and second row, we reshape the data from $N_x/N_{filters} \times N_y \times 
N_{filters}$ to $N_x \times N_y \times 1$, thereby transferring the information from the 
filters dimension into the spatial $x$ dimension. Each 
layer is followed by an activation function. We use a linear rectifier $x\rightarrow 
\max(0,x)$, while for the last layer we use the sigmoid function $x\rightarrow \frac{1}{1+e^{-
x}}$ in order to obtain an image of probabilities (indeed, the sigmoid function is 
frequently used for segmentation problems).

The parameter vector $\theta$ consists of the weights $W_i$ and biases $b_i$ of the fully-
connected layers $\sigma(A)_i = W_i A + b_i^T$, as well as  the weights of the convolutional 
filters $w_i$ corresponding to the convolutional layers $\sigma(A)_i = w_i * A + b_i^T$.

The loss function we use for training the network, the target of the Stochastic Gradient 
Descent (SGD) algorithm, is the Negative Log-Likelihood (NLL) computed over the entire training 
set:
\begin{equation}\label{eq:nll-loss-def}
loss_{NLL}(\theta)\equiv \ell_{NLL}(\theta) = \frac{1}{N_{train}}\sum_{q=1}^{N_{train}} \ell_{CE}(\widetilde{\IM}(d_q;\theta),\IM_q),
\end{equation}
where $\{\IM_q\}_{q=1}^{N_{train}}$ are the true train labels as defined in \eqref{eq:plateau-image-def}, and $\ell_{CE}$ is the pixel-wise cross-entropy loss
\begin{equation*}
\ell_{CE}(\widetilde{\IM},\IM)=-\frac{1}{N_x \cdot N_y} \sum_{i=1}^{N_x}\sum_{j=1}^{N_y}(\IM)_{i,j}\log(\widetilde{\IM})_{i,j}.
\end{equation*}

Informally, the NLL loss approximates the total likelihood of pixels from the prediction to be of similar value as the ones in the true label. Minimizing $\ell_{CE}$ is equivalent to maximizing the overlap between the predicted plateaus and the true plateaus. Since $\tilde{\IM}\neq 0$ (it is the output of the sigmoid function in the last layer), $\ell_{CE}$ (and therefore $\ell_{NLL}$) is clearly differentiable.

\subsection{Physically Informed loss}\label{sub:pi-loss}
We have found by experimentation (see \Cref{results}) that the network trained with minimizing the $\ell_{NLL}$ loss was performing well on its own. However, the $NLL$ loss does not take into account any knowledge of the underlying physical problem. We implement a \textit{physically-informed} loss term by using the true and the predicted images $\IM$, $\tilde{\IM}$ as (discretized) \emph{source functions} in the Helmholtz equation, and comparing the pressure fields which would result from those sources in the search domain. By utilizing the physical knowledge of the problem, we expect better performance (e.g. convergence speed and accuracy) of the network.

In more detail, let $f(\xb)$ and $\tilde{f}(\xb)$ denote some functions $f,\tilde{f}:\mathcal{S} \to [0,1]$ such that $\IM,\tilde{\IM}$ are precisely $f,\tilde{f}$ sampled on the $N_x\times N_y$ discretization ${\cal{S}}_d$ of the search domain $\mathcal{S}$. Considering these $f,\tilde{f}$ as \emph{extended sources}, the solution of the corresponding Helmholtz equation can be written as
\begin{equation} \label{eq:field_ext}
p(\w,\xb) = \int_{\mathcal{S}} G(\xb,\yb;\w) f(\yb)~\rmd \yb, \qquad \tilde{p}(\w,\xb)=\int_{\mathcal{S}} G(\xb,\yb;\w) \tilde{f}(\yb)~\rmd \yb,
\end{equation}
where $G$ is the Green's function as shown in \cref{eq:greensf}.
Since $\tilde{f}$ is available only as the network output $\tilde{\IM}(d;\theta)$, we compute an approximation to $\tilde{p}$ in any location $\xb\in\mathcal{S}$ by the quadrature formula
\begin{equation}\label{eq:quad-approx-to-ptilde}
\tilde{p}(\w,\xb) \approx \sum_{\yb_j\in {\cal{S}}_d} G(\xb,\yb_j;\w) \left(\tilde{\IM}(d;\theta)\right)_{\yb_j}=:\tilde{V}(\xb;d,\theta),\quad \xb\in\mathcal{S}.
\end{equation}

For simplicity, we use the same method to approximate the true pressure field $p$ generated by the extended source $f$, i.e.
\begin{equation}\label{eq:quad-approx-to-p}
p(\w,\xb) \approx \sum_{\yb_j\in {\cal{S}}_d} G(\xb,\yb_j;\w) \left(\IM\right)_{\yb_j}=:V(\xb;\IM),\quad \xb\in\mathcal{S}.
\end{equation}

Finally, using \cref{eq:quad-approx-to-p} and \cref{eq:quad-approx-to-ptilde} we compute an approximation to the discrepancy between the pressure fields $p^q,\tilde{p}^q$ (corresponding to the sources $\{\xb_{i,q}\}_{i=1}^{N_q}$) over the entire training set as 
\begin{equation*}
\ell_{PI}(\theta) = \frac{1}{N_{train}\cdot N_x N_y}\sqrt{\sum_{q=1}^{N_{train}}\sum_{\xb\in\mathcal{S}_d}\left| V(\xb;\IM_q) - \tilde{V}(\xb;d_q,\theta) \right|^2}  \approx \|p-\tilde{p}\|_2.
\end{equation*}

As we will see in the next section, this new physically-informed loss term helps the network converge to the desired solution more effectively.
 We train the network using both NLL and physically-informed loss term such that
 \begin{equation}\label{eq:PI-loss-def}
 loss_{PI}(\theta) = 0.5\cdot \ell_{NLL}(\theta) + 0.5\cdot \ell_{PI}(\theta)
 \end{equation}
 and in this setup we achieved the best results. If one were to use only the PI term $\ell_{PI}$ as the loss function, the network would not converge. This is due to the random initialization of the weights, causing the outputs to be random as well. This ``pure'' PI loss, which is based on propagating a field generated by approximated sources, would saturate on a local minimum. On the other hand, the robustness of the NLL network allows us to generate accurate predictions of the sources, limited to the information in the data set. Using the knowledge of the underlying physical problem, the $\ell_{PI}$ term acts as a regularizer, penalizing the network on bad predictions. By combining $\ell_{NLL}$ and $\ell_{PI}$, we achieve the best performance. The choice of averaging the PI and NLL terms is justified by the fact that both losses are of the same order of magnitude.

\section{Numerical Results} \label{results}
In this section, we present how the method performs and  assess the effectiveness of the PI loss on the validation loss of the network.
\subsection{Setup}
We assume a homogeneous waveguide with constant wave 
speed  $c_0=1500$~m/s and depth $D=200$~m. Homogeneous Dirichlet boundary conditions are applied on the 
top and bottom boundaries. The sources inside the waveguide emit signals across multiple frequencies and 
specifically have a central frequency $f_c=32.0625$~Hz and a bandwidth $B=0.4f_c$, i.e. the emitted 
frequencies are in the range $f\in \left[ f_c - B/2, f_c + B/2\right]$. We uniformly discretize the frequency interval 
such that we use a total of $N_f= 33$ frequencies. Furthermore, the vertical array is fixed at $x=0$ and 
spans the entire depth of the waveguide, with an inter-element distance of $h = 2.5$~m, resulting in $N_r = 
81$.

We choose a search domain $\mathcal{S} = [490, 570] \times [0, D]$~m, which we discretize with a square 
grid with $h_x = h_y = 4$~m, thus $N_x = 71$ and $N_y = 51$. We generate $N_{samples} = 5000$ samples, out of which
$N_{train}=4050$ were used for training, 450 were used for validation, while the remaining 500 were used for 
testing. For each sample, the number of sources is randomly generated to be between 1 and 6, while their locations are also randomized per sample. Lastly, the network labels are created with an $N_p \times N_p$ plateaus around each source location as in \cref{eq:plateau-image-def}, with $N_p=3$.

\subsection{Training}
We trained the network using the stochastic gradient descent (SGD) algorithm ADAM 
\cite{adam}, with 50 epochs and a batch size of $8$. We use the validation set to prevent the 
network from over-fitting the train data. If, during training, the computed loss over the 
training data continues to decrease and the computed loss over the validation data increases 
(using the same loss function), it means that the network is over-learning the training data 
and loses the ability to generalize to samples outside of the train data. Before we observe 
over-fitting, we terminate the training process. 

\subsection{Evaluation}\label{eval_method}

To evaluate the performance of the trained network, we calculate the number of sources that the model located successfully out of all the sources in the testing data set. We first locate the sources in the output images by using a \textit{mean filter}, where we replace each pixel value in an image with the mean value of its neighbors, including itself. This operation can be described by a convolution with a filter whose kernel is precisely $\frac{1}{N_p^2} \PL_{N_p} $ as in \cref{eq:plateau-image-def}.

\begin{figure}[ht]
\begin{center}
\includegraphics[width=0.95\linewidth,clip]{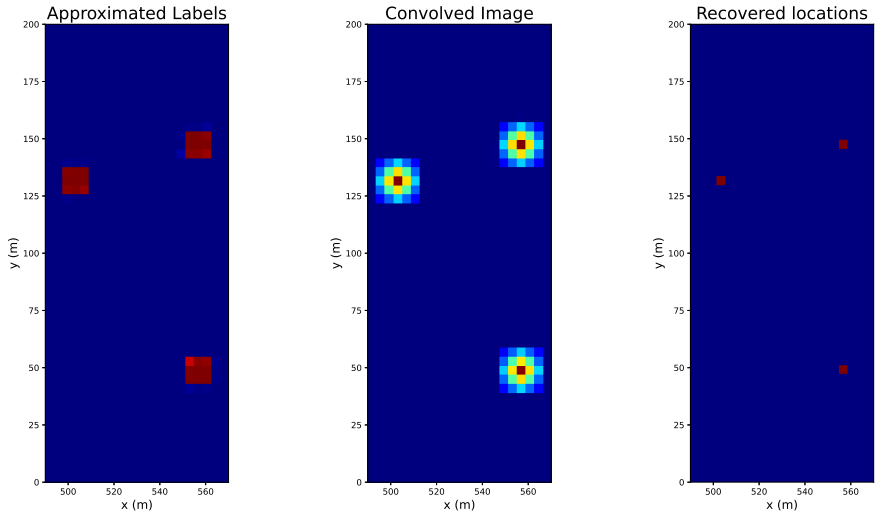}
\caption{Approximated labels, generated by the network (left), Mean Filter version of the labels (middle) and recovered locations of the sources (right).}
\label{fig:eval_method_conv}
\end{center}
\end{figure}

An example of this evaluation process is shown in \Cref{fig:eval_method_conv}. In the leftmost image, we plot the values of the approximates labels generated by the network for a single sample. In the middle plot, we show the effect of the mean filter in the image. We observe that this smoothed version of the image exhibits peaks in the locations of the point sources. From this, it is easy to recover the approximated locations of the sources, by simply thresholding the image for values close to 1. The result of this thresholding is shown in the rightmost plot, which shows the recovered approximate locations of the sources in this sample.

 For each source, we check if the network managed to locate it precisely, by comparing it with the corresponding image that contains the true locations of each source in that sample. It is often sufficient to seek the source in a small area around the true location, however we chose to compare the exact locations, since in this case we are testing the resolution problem in the most challenging fashion.

\subsection{Results}

\begin{figure}[ht]
\begin{center}
\includegraphics[width=0.95\linewidth,clip]{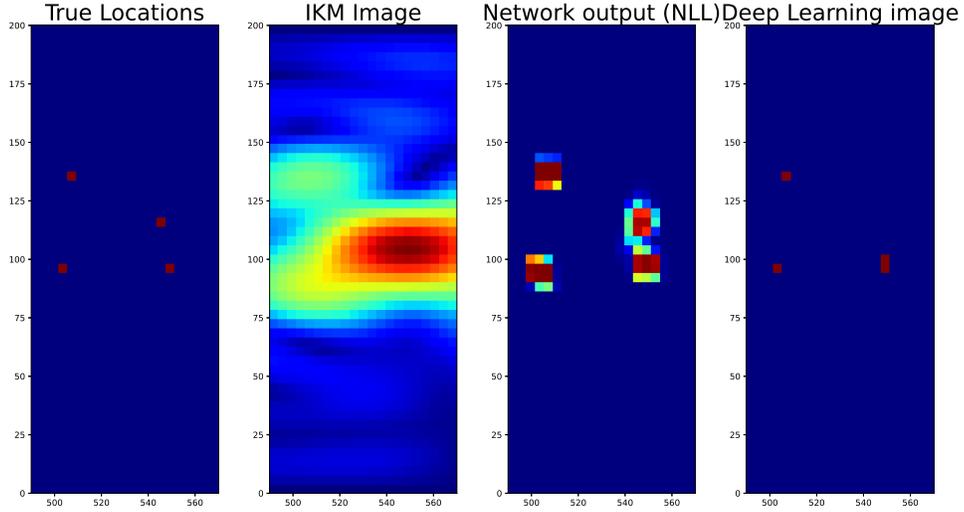}
\caption{From left to right: True locations of the sources, $\IKM$ image, network output when only using the NLL loss.}
\label{fig:num_exp_comp_NLL}
\end{center}
\end{figure}

In \Cref{fig:num_exp_comp_NLL} we present a visual example of the network output for the case where only the NLL loss $loss_{NLL}$ in \eqref{eq:nll-loss-def} is used. In the first plot, we show the true locations of the sources we wish to image. In the second column we plot the modulus of the $\IKM$ image, given by \cref{eq:IKM}. The traditional imaging functional has trouble separating the sources effectively, as the resulting image focuses on the two rightmost sources that are also the ones closer together. Next, we plot the output of the network, i.e. the approximated labels. We see that while the approximate locations of the sources are recovered well, the overall image is distorted. Lastly, when trying to recover the exact locations of the sources, by using the evaluation method we just described. we can see that the network has managed to locate the two sources on the left that are well-separated, but did not manage to locate the sources that were placed closer to each other.

\begin{figure}[ht]
\begin{center}
\includegraphics[width=0.95\linewidth,clip]{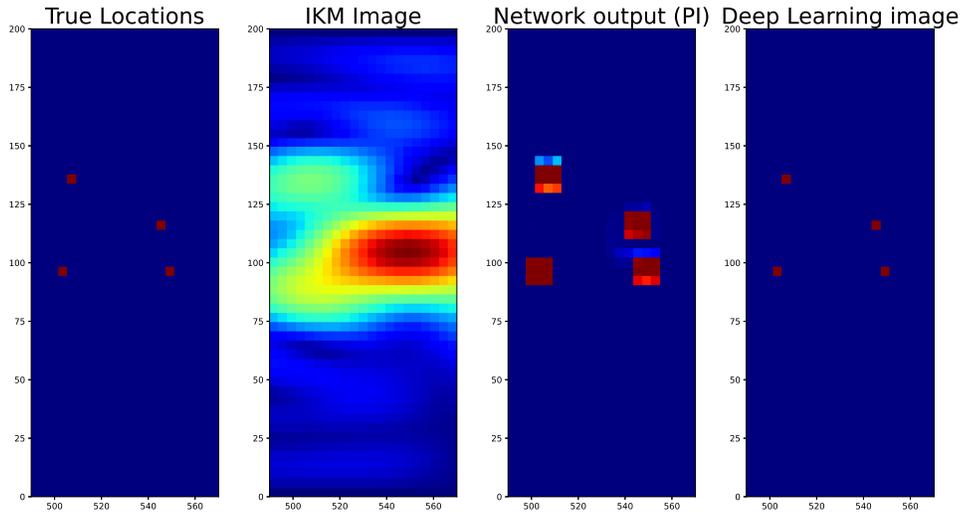}
\caption{From left to right: True locations of the sources, $\IKM$ image, network output when only using  the physically-informed loss.}
\label{fig:num_exp_comp_PI}
\end{center}
\end{figure}

Next, in \Cref{fig:num_exp_comp_PI} we repeat the same experiment as in
\Cref{fig:num_exp_comp_NLL}, where we now use $loss_{PI}$ in \eqref{eq:PI-loss-def} as the loss function during training. In the last two plots we see that the generated labels, plotted in the third column have vastly improved in quality. In the last plot, this translates to the perfect recovery of all the sources, including the two close-by sources that were place on the right.

\begin{figure}[ht]
\begin{center}
\includegraphics[width=0.95\linewidth,clip]{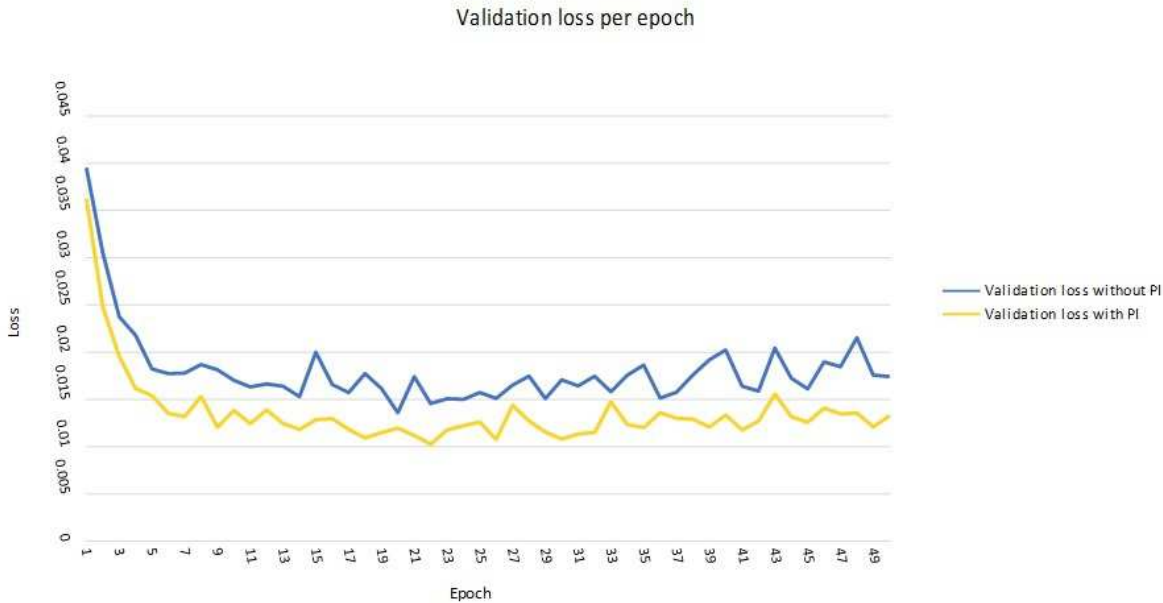}
\caption{Validation loss per epoch, without the usage of the physically-informed loss term (blue) and with the implementation of the physically-informed loss term (yellow).}
\label{fig:loss_with_without_PI}
\end{center}
\end{figure}

Having seen the qualitative improvement that the physically-informed loss offers to the model, we want to also examine its performance in a quantitative way. In  \Cref{fig:loss_with_without_PI} we plot the validation loss of the network with respect to the epochs, and we observe that the validation loss corresponding to the usage of the physically-informed loss, represented by the yellow line, is immediately smaller than if that loss term was not used, represented by the blue line. Moreover, the recovery rate for the source locations starts at $59\%$ when using the NLL loss, but jumps to $75\%$ when using the physically-informed loss, thus showing the major contribution of the knowledge of the physical problem to the accuracy of the proposed method.

\begin{figure}[ht]
\begin{center}
\includegraphics[width=0.95\linewidth,clip]{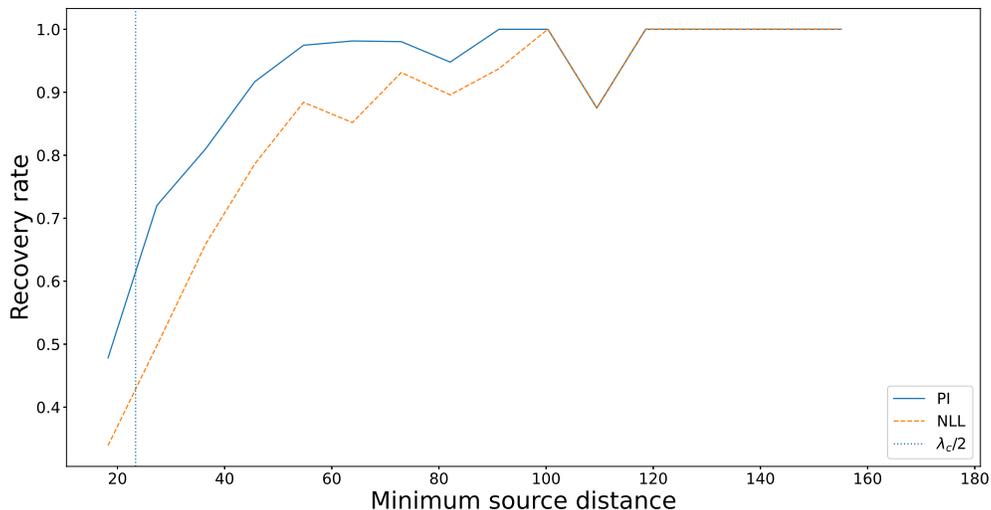}
\caption{Recovery rates of sources, as a function of the minimum distance between sources per sample.}
\label{fig:sources_rec_min_dist}
\end{center}
\end{figure}

While  \Cref{fig:loss_with_without_PI} clearly indicates the superior performance of the network when using the PI loss, we would still like to have more insight into the benefits of this situation.  In \Cref{fig:sources_rec_min_dist} we plot the recovery rates of the sources, with respect to the minimal distance of sources in each sample, i.e. 
$$\min_{\begin{array}{c} \scriptstyle i,j=1,2,\ldots, N_s,\\ \scriptstyle i \neq j
\end{array} } \left|\xb_i - \xb_j \right|.$$
We see that the PI loss, plotted with a solid blue line, consistently outperforms the NLL loss, plotted with a dashed red line. It is of exceptional interest that the discrepancy between the two methods is the largest near the Rayleigh limit $\lambda_c/2$, plotted with a vertical dotted line. It is of great interest that in the super-resolution regime, where the minimum distance is below $
\lambda_c/2$, the PI loss massively outperforms the NLL loss, with the difference in performance between the two methods becoming less significant as we move to higher minimum distances.

Finally, we examine the performance of our model in the presence of measurement noise. 
Specifically, we use the network trained on noiseless data, on test data which is contaminated by noise. The goal is to study how different levels of noise affect the ability of the model to recover the sources. Each clean test sample $d_q\in \mathbb{C}^{N_r\times N_f}$ as in \eqref{eq:exact-data} is modified as follows:
\begin{equation}\label{eq:noisy-sample}
\tilde{d}_q(r,j) = d_q(r,j) + W_q(r,j), \quad r=1,\dots,N_r,\;j=1,\dots,N_f,
\end{equation}
where $W_q(r,j)$ is chosen according to some distribution. Two noise distributions were tested: uniform and Gaussian.

To implement the uniformly distributed noise, we set for each $d_q$ in the test set (here $\mathcal{U}([a,b])$ denotes the uniform distribution over the interval $[a,b]$)
\begin{align*}
W_q(r,j)&=\Re{d_q(r,j)}\cdot(1+w_{1,q}(r,j))+\imath \Im{d_q(r,j)}\cdot(1+w_{2,q}(r,j))\\
 w_{1,q}(r,j),w_{2,q}(r,j) &\sim \varepsilon \mathcal{U}
 ([-1/2,1/2]),\quad
r=1,\dots,N_r,\;j=1,\dots,N_f.
\end{align*}

\begin{figure}[ht]
\begin{center}
\includegraphics[width=0.95\linewidth,clip]{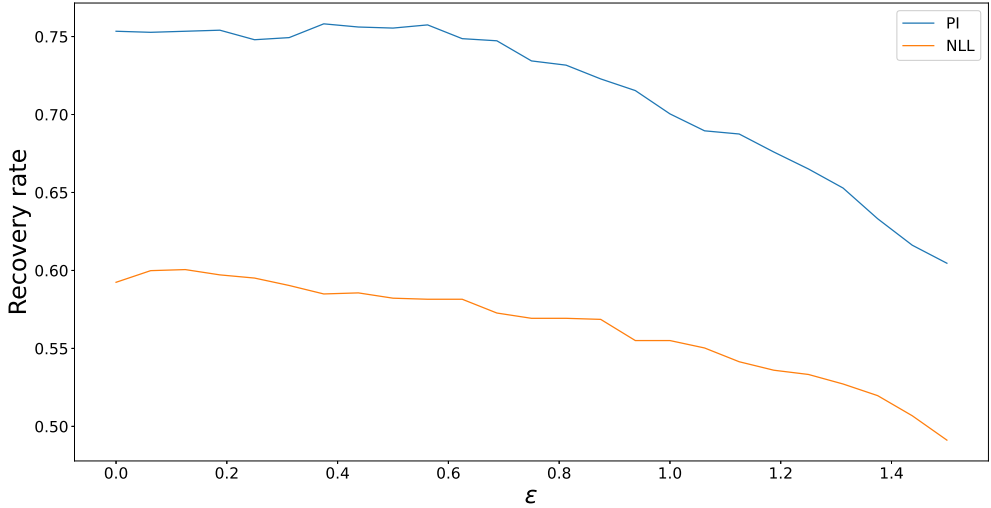}
\caption{Recovery rates of sources with uniformly perturbed test data.}
\label{fig:sources_rec_noises_uniform}
\end{center}
\end{figure}

Thus, both real and imaginary parts of the data are perturbed by uniform noise of relative magnitude $\varepsilon$. The performance of the network as a function of $\varepsilon$ is shown in \Cref{fig:sources_rec_noises_uniform}. The model is extremely stable both for the NLL and the PI loss.

For the Gaussian noise, we follow \cite{BPV_08}. For each test sample $d_q$ and $j=1,\ldots,N_f$, denote by $d_q^{(j)}$ the $j$-th column of the data matrix, corresponding to the measurements obtained by the entire receiver array at the frequency $\omega_j$. For $\varepsilon>0$, let  $w_q^{(j)}\in\mathbb{C}^{N_r}$ be a zero mean uncorrelated Gaussian distributed vector with variance $\varepsilon p_{avg}$, \
where 
$$p_{avg} = \frac{1}{N_r} \left\|d_q^{(j)} \right\|^2_2,$$
i.e. $w_q^{(j)}(r) \sim \mathcal{N} (0, \varepsilon p_{avg})$. Finally, the noise matrix in \eqref{eq:noisy-sample} is 
$$
W_q=[w_q^{(1)}\;\dots\;w_q^{(N_r)}].
$$

Thus, the expected power of the noise for the measurement at frequency $\omega_j$
 over all the receivers is  
$$\mathbb{E}\left[ \left\|w_q^{(j)} \right\|^2_2 \right] = \varepsilon N_r p_{avg}, $$ 
while  the total power of the signal recorded on all the receivers is $N_r p_{avg}$. Therefore, 
our Signal-to-Noise Ratio (SNR) in dB is given by $-10 \log_{10} \varepsilon$.

\begin{figure}[ht]
\begin{center}
\includegraphics[width=0.95\linewidth,clip]{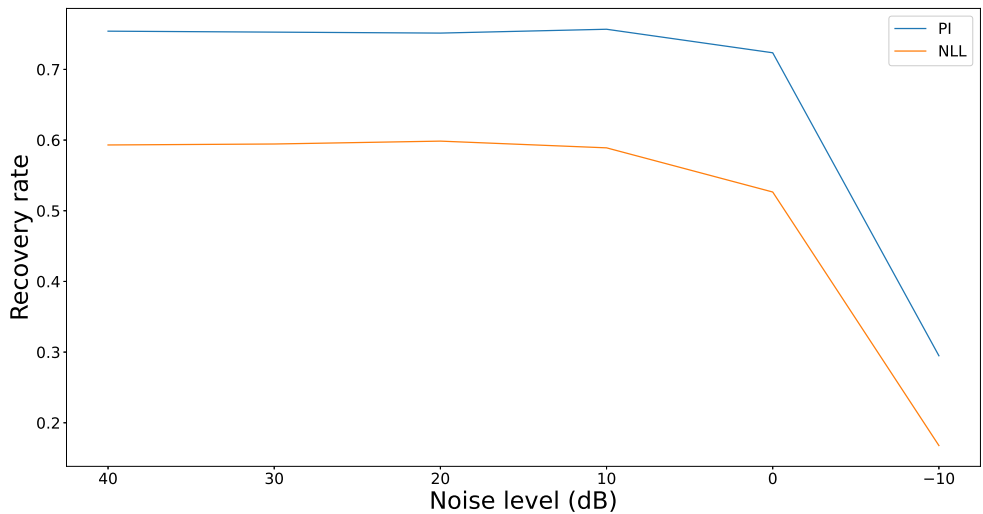}
\caption{Recovery rates of sources with Gaussian noise.}
\label{fig:sources_rec_noises_gaussian}
\end{center}
\end{figure}

In \Cref{fig:sources_rec_noises_gaussian}, we plot the recovery rates of our model, under 
varying levels of Gaussian measurement noise. We have chosen $\varepsilon= 10^{-4},10^{-3},
10^{-2},0.1,1,10$, which correspond to SNR values of $40,30,20,10,0,-10$~dB respectively.  As 
can be immediately seen, our model performs very well under the presence of noise, only losing 
accuracy for the first time at an SNR of 0~dB. For the sake of clarity, let us note here that 
a choice of $\varepsilon=1$, which corresponds to an SNR value of 0, the average power of the 
measurement noise is the same as the average power of the signal. Our last point at $-10$~dB 
sees a sharp drop in performance, but that is to be expected, as at this point the power of the 
noise far exceeds the power of the recorded signal.

\section{Summary and conclusions}

In this work we have demonstrated a successful deep learning method to waveguide imaging of multiple point sources, overcoming the Rayleigh resolution limit in practice. We have also shown that adding a physically informed loss term results in significant improvement in the accuracy of the network predictive power. Future work includes imaging in more complicated  geometries, in particular where there is no known analytical solution (such as waveguides with variable depth, terminating and inhomogeneous waveguides), and a 3D implementation. Moreover, it will be interesting to compare the approach to other emerging DL methods to super-resolution, e.g. \cite{li_accurate_2021}.

\bibliographystyle{unsrt}  
\bibliography{references}

\end{document}